\documentclass[sigconf]{acmart}
\usepackage{multirow}
\usepackage{xcolor}
\usepackage{adjustbox}

\usepackage[toc,page]{appendix}
\usepackage{caption}
\usepackage{comment}
\usepackage{subcaption}
\usepackage{enumitem}
\usepackage{hyperref}

\newcommand{\cocowidth}{0.23} 

\usepackage{xspace}
\makeatletter
\DeclareRobustCommand\onedot{\futurelet\@let@token\@onedot}
\def\@onedot{\ifx\@let@token.\else.\null\fi\xspace}

\def\eg{\emph{e.g}\onedot}

\def\etc{\emph{etc}\onedot} 
 
\def\etal{\emph{et al}\onedot}
\makeatother

\AtBeginDocument{%
  \providecommand\BibTeX{{%
    \normalfont B\kern-0.5em{\scshape i\kern-0.25em b}\kern-0.8em\TeX}}}

\acmConference[Technical report]{2021}{2021}

\begin{document}

\title{HIME: Efficient Headshot Image Super-Resolution with Multiple Exemplars}


\author{Xiaoyu Xiang$^{1,2}$, Jon Morton$^1$, Fitsum A Reda$^{1,\dagger}$, Lucas Young$^1$, Federico Perazzi$^{1,\dagger}$, Rakesh Ranjan$^1$, Amit Kumar$^1$, Andrea Colaco$^{1,\dagger}$, Jan Allebach$^2$}
\affiliation{%
  \institution{$^1$Meta Reality Labs, $^2$Purdue University}
  \country{}
}

\renewcommand{\shortauthors}{Xiang, et al.}

\begin{abstract}
\let\thefootnote\relax\footnote{$^{\dagger}$Affiliated with Meta at the time of this work.
.}
A promising direction for recovering the lost information in low-resolution headshot images is utilizing a set of high-resolution exemplars from the same identity. Complementary images in the reference set can improve the generated headshot quality across many different views and poses. However, it is challenging to make the best use of multiple exemplars: the quality and alignment of each exemplar cannot be guaranteed. Using low-quality and mismatched images as references will impair the output results. To overcome these issues, we propose an efficient \textbf{H}eadshot \textbf{I}mage Super-Resolution with \textbf{M}ultiple \textbf{E}xemplars network (HIME) method. Compared with previous methods, our network can effectively handle the misalignment between the input and the reference without requiring facial priors and learn the aggregated reference set representation in an end-to-end manner. Furthermore, to reconstruct more detailed facial features, we propose a correlation loss that provides a rich representation of the local texture in a controllable spatial range.
Experimental results demonstrate that the proposed framework not only has significantly fewer computation cost than recent exemplar-guided methods but also achieves better qualitative and quantitative performance.
\end{abstract}




\maketitle
\section{Introduction}
\label{sec:intro}

Numerous psychological and cognitive studies have shown that face perception is one of the most important and specialized aspects of social cognition~\cite{farah1998special,quinn2011face}. 
The facial regions of a picture tend to draw the attention and interest of observers immediately. Moreover, humans are susceptible to minor changes in familiar faces~\cite{rhodes2011oxford}. Thus, increasing the quality of the face region in images and videos has the potential to significantly enhance the user experience of many social communication applications, \eg real-time video chat, mobile photo booth, \etc.

\begin{figure}[tbp]
    \centering
    \includegraphics[width=0.9\linewidth]{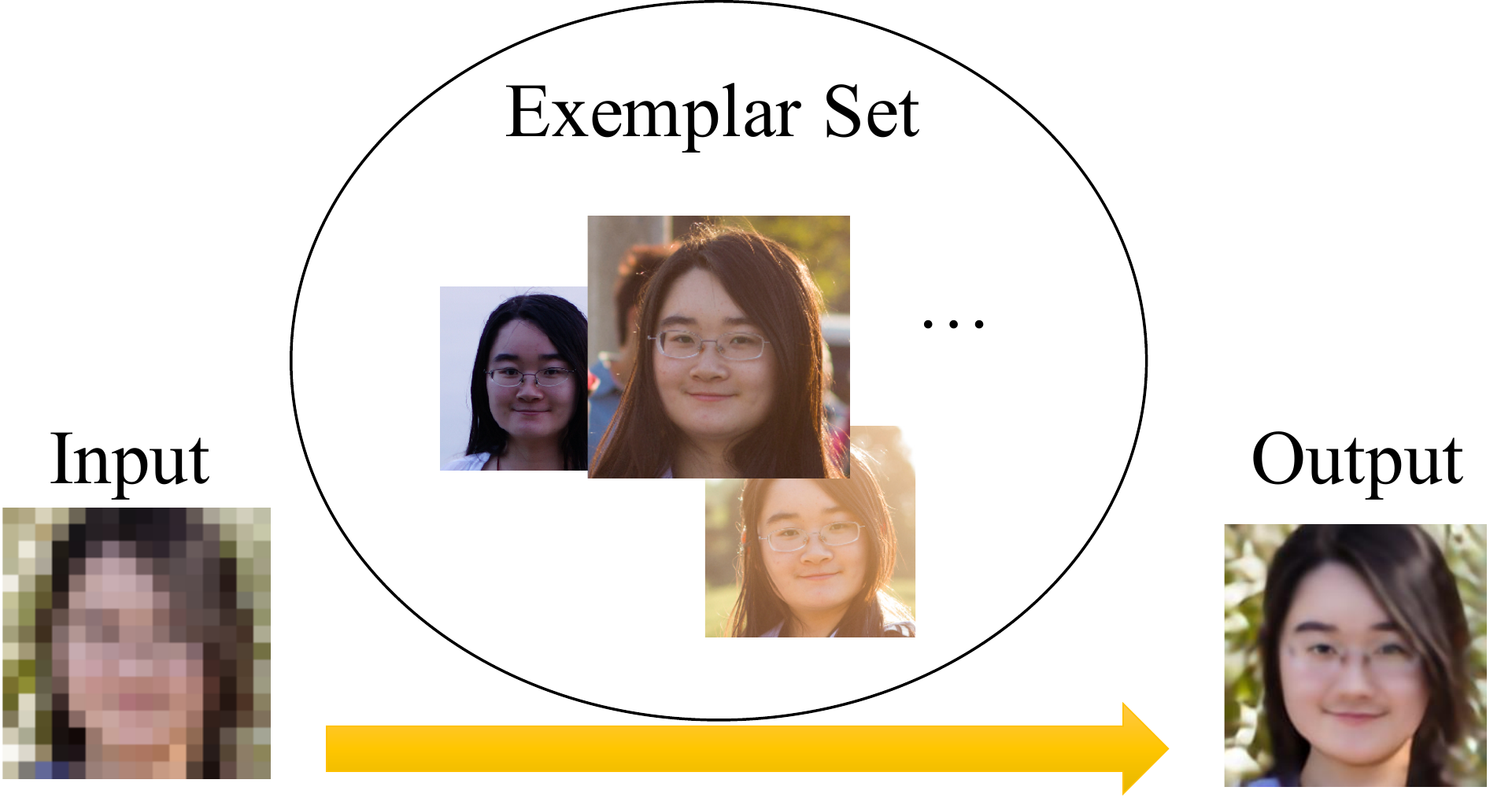}
    \caption{Headshot super-resolution that recovers the lost information in the input using a set of exemplars.}
    \label{fig:teaser}
    \vspace{-2mm}
\end{figure}

For the above reasons, the machine learning community has widely explored face hallucination~\cite{park2008example,yang2013structured,song2017learning,cao2017attention,chen2018fsrnet,bulat2018learn} as a domain-specific problem of single image super-resolution(SISR)~\cite{allebach1996edge,atkins2001optimal,xiang2020boosting}, which aims to restore realistic details from a low-resolution~(LR) face image to a high-resolution (HR) one. Benefiting from the integration of face structure and identity priors and recent progress in deep neural network designs, it is now possible to generate visually pleasing results even for extremely tiny faces. When the input LR headshot does not contain enough attribute or identity information, using additional references can help to achieve a more faithful reconstruction result. In this paper, we explore a novel method that makes full use of an arbitrarily-sized set of exemplar images to increase the fidelity of headshot image super-resolution. 

One core problem is to search the matching regions from references and transfer the corresponding features to the output. Previous methods choose to conduct the global context matching with registration~\cite{yue2013landmark}, optical flow~\cite{wang2016light,zheng2018crossnet,li2018learning,dogan2019exemplar} with a warping~\cite{shim2020robust}. Still, these works assume the exemplars share a similar viewpoint with the LR input~\cite{shim2020robust}, which cannot always be guaranteed. Besides, their performance depends on accurate motion estimation and may poorly capture long-range correlations. Other methods~\cite{boominathan2014improving, zhang2019image,yang2020learning,xie2020feature} conduct an exhaustive patch-wise comparison of LR and reference features, which require a large amount of computation, especially when the reference resolution is high. In addition, these methods cannot handle inter-patch misalignment or non-rigid deformations. To better use the information of faces from different poses or views, we propose a Reference Feature Alignment module (RFA) that combines optical flow and deformable alignment to find the corresponding information in reference features and align them with the LR content inspired by~\cite{lin2021fdan,cai2021space,chan2021basicvsr++}.

In practical applications like smart home cameras or mobile photography, it is possible to acquire many high-resolution images of different views when the user is close to the camera. These images can naturally serve as good exemplars to enhance far-away tiny faces. However, most previous works focus on reference-based super-resolution (RefSR) with one exemplar~\cite{zhang2019image,yang2020learning,shim2020robust,li2018learning,dogan2019exemplar}, which is a simplified assumption. To handle a set of exemplars, these methods require an extra step to select the most similar image as the reference according to SIFT~\cite{lowe1999object,zhang2019texture} or facial landmarks points~\cite{li2020enhanced}, which is a poor representation of the whole set. \cite{wang2020multiple} devises a framework to process and combine multi-exemplars with a weighted pixel average. Still, it is not robust to the displacement or distortions in reference images, as is our method. To utilize the reference set effectively and efficiently, we propose a Content-conditioned Feature Aggregation module (CoFA) that simplifies the set-to-image RefSR problem to a point-to-point RefSR by aggregating feature maps in a set into a single representation. 

Benefiting from the module designs above, our network is end-to-end trainable without requiring other face-specific meta-information. Aiming to generate an SR output with highly-detailed textures, we propose a novel correlation loss inspired by the correlation layer in FlowNet2~\cite{ilg2017flownet,flownet2-pytorch} to supervise the reconstruction of texture patterns. We compute the pixel-wise correlation across the channel dimension to represent the local textures within a certain window size.

In summary, our contribution is four-fold: (1) we propose a novel headshot super-resolution network that takes advantage of multiple exemplars. Our method is more effective than previous approaches by thoroughly integrating the corresponding information in the exemplar set. It is also computationally efficient since we conduct the matching and transferring in the LR space with careful design; (2) we propose a novel reference feature alignment network to find and align corresponding reference features to the LR content based on flow-guided deformable sampling. We devise a feature aggregation module conditioned on the LR content to explicitly improve the set representation by favoring features that are high in quality and similarity; (3) we propose a novel correlation loss that helps represent the local texture and reconstruct more realistic details; (4) compared with previous approaches, our method achieves state-of-the-art face hallucination performance on the CelebAMask-HQ testset. It also has fewer parameters and computational costs than recent exemplar-guided methods.

\section{Related Works}
\label{sec:related_works}
\subsection{Reference-based Super-Resolution}
Reference-based SR (RefSR)~\cite{freeman2002example} can reconstruct more accurate structures and details benefiting from the reference HR image. The general solution of RefSR includes two steps: searching the matched textures between LR inputs and HR references, and transferring the textures. Some of the previous RefSR approaches choose to align the LR and Ref images with either global registration~\cite{yue2013landmark} or optical flow~\cite{wang2016light,zheng2018crossnet}. Other methods choose to match by patches with gradient features~\cite{boominathan2014improving}, or deep features extracted by the CNN~\cite{zhang2019image,yang2020learning,xie2020feature}. \cite{shim2020robust} change the feature matching to LR space to reduce computation. \cite{yang2020learning} introduced the transformer architecture in a cross-scale manner to improve the accuracy of searching and transferring relevant textures. The above works usually include pixel-wise reconstruction loss, perceptual loss~\cite{simonyan2014very} and adversarial loss as the objective functions. Zhang \etal~\cite{zhang2019texture} introduce a Haar wavelet loss and a degradation loss to avoid over-smoothing in final results. 
Besides, CMSR~\cite{cui2020towards} further expands the reference source from a single image to a pre-built image pool and searches the $k-$nearest patches from the pool. Since these methods exhaustively conduct a patch-wise comparison of LR and reference feature maps, they usually have a high computational cost.

\subsection{Face Hallucination}
Face hallucination methods can be roughly divided into two categories: blind face hallucination and exemplar-guided restoration. The first category focuses more on integrating face priors in designing the reconstruction network and loss functions: some works include sub-branches for facial landmarks or face structures
~\cite{zhu2016deep,song2017learning,bulat2018super,yu2018face,kim2019progressive,yin2020joint}, or face parsing map~\cite{chen2018fsrnet,chen2020progressive}. Using face structure priors may bring advantages, including the better recovery of the face shape, as reflected by fewer errors on face alignment and parsing. However, the reconstruction results might not look like the same person, especially when the input images contain barely any identifying information. To solve this problem, \cite{zhang2018super,hsu2019sigan,grm2019face} employ identity information to supervise the training of the reconstruction network. However, these blind reconstruction methods are heavily influenced by the bias within the distribution of training data, and usually fail to generate satisfying results for minority groups.

\begin{figure*}[t]
    \centering
    \includegraphics[width=\linewidth]{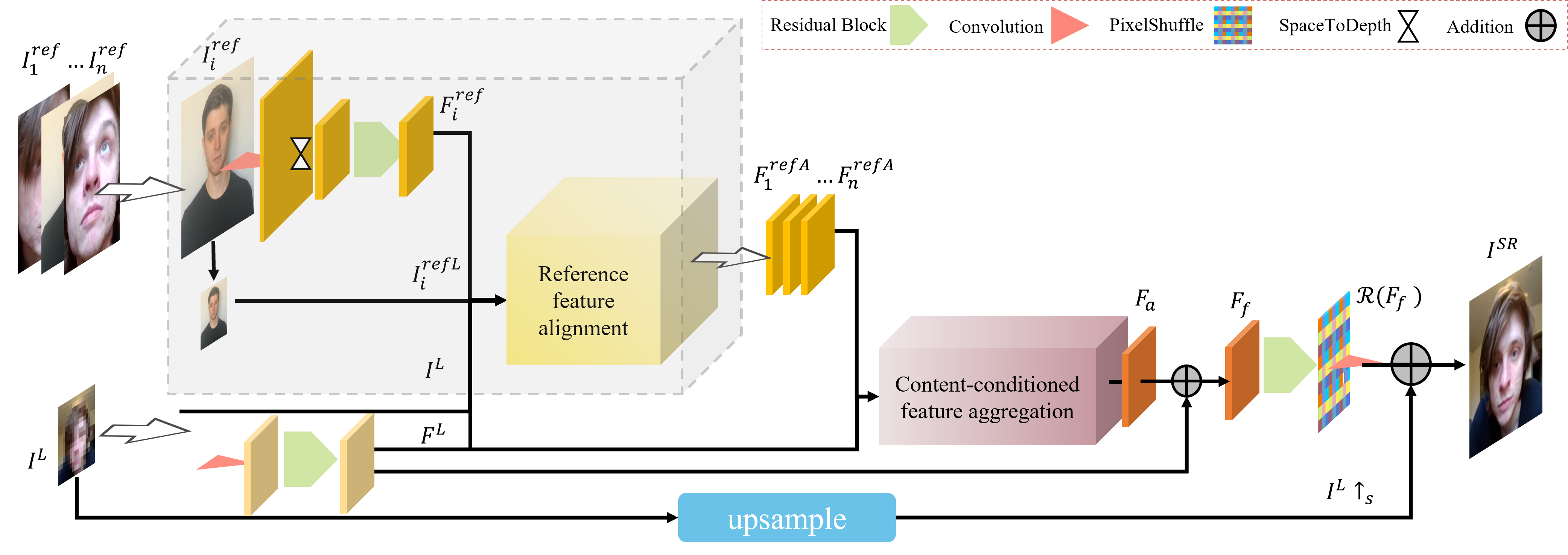}
    \vspace{-2mm}
    \caption{Overview of our \textbf{H}eadshot \textbf{I}mage Super-Resolution with \textbf{M}ultiple \textbf{E}xemplars (HIME) framework. Given an input LR image and any number of exemplars, it matches, aligns, and aggregates the features of the reference images conditioned on the input content to reconstruct the SR output. }
    \label{fig:framework}
    \vspace{-3mm}
\end{figure*}

The second category, exemplar-guided restoration, aims to use another HR image of the same person to improve the visual content quality of the generated images. \cite{li2018learning,dogan2019exemplar} include a warping sub-network in using the HR guidance, which increases the training steps as well as the computation cost of the network. \cite{li2020enhanced} uses moving least-squares to align the input and guidance images in the feature space and applies AdaIN for feature transfer. It selects a single exemplar from the guidance images, thus cannot fully use the rich information in the guidance face sets. \cite{wang2020multiple} takes a step forward by using multiple exemplars with a weighted pixel average module in the network. But, it cannot handle the large deformation between unaligned faces. 

Compared with the approaches above, our method can take full advantage of an unaligned exemplar set as a reference in headshot reconstruction, and our network is end-to-end trainable without requiring face-specific metadata.

\section{HIME Framework}
\label{sec:framework}

Given a low-resolution input $I^{L}$ and a set of high-resolution headshot images $\mathcal{I}^{ref}=\{I^{ref}_i\}, i=1,2,\ldots$ from the same identity, our goal is to generate the corresponding high-resolution image $I^{SR}$. To efficiently and accurately transfer the matching information from the unaligned reference sets of arbitrary length, we propose the \textit{HIME} framework as illustrated in Figure~\ref{fig:framework}. This framework consists of four main components: \textit{feature extractor}, \textit{reference feature alignment module (RFA)}, \textit{content-conditioned feature aggregation module (CoFA)}, and \textit{HR reconstructor}, as introduced in Sections~\ref{subsec:fea_ext}, \ref{subsec:rfa_module}, \ref{subsec:cofa_module} and \ref{subsec:hr_recon}.


We first use an LR feature extractor to get the feature map $F^{L}$ from $I^L$ and an HR feature extractor to get feature maps $\{ F^{ref}_i\}_{i=1}^n$ from the reference set with $n$ HR images. For efficient feature matching and transfer, the reference images and features are converted to the LR space. Then we feed $I^L$, $F^L$, $\{ I^{ref}_i\}_{i=1}^n$ and $\{ F^{ref}_i\}_{i=1}^n$ to the proposed RFA module for alignment. Furthermore, to better utilize the face set information, we use a CoFA module to aggregate the refined features into one. Finally, we reconstruct the HR face image from the aggregated feature map.

\subsection{Feature Extractors}
\label{subsec:fea_ext}
We adopt an HR feature extractor and an LR feature extractor to handle images in HR space and LR space, respectively. The HR feature extractor turns the HR reference images into a set of feature maps: $\{ F^{ref}_i\}_{i=1}^n$. The RGB images are first converted into a mono-channel feature map since the color information of the reference images is not needed. Then, we adopt a space-to-depth operation to convert the HR feature maps into the same spatial resolution as the input without discarding any information. Next, we apply a convolution layer and $k_h$ residual blocks~\cite{he2016deep} to extract the HR reference feature maps. The LR feature extractor generates feature maps for the input LR image with a convolutional layer and $k_l$ residual blocks~\cite{he2016deep}.

\subsection{Reference Feature Alignment}
\label{subsec:rfa_module}
\begin{figure}[htbp]
\captionsetup[subfigure]{labelformat=empty}
\begin{center}
  \begin{subfigure}[b]{\linewidth}
  \includegraphics[width=\linewidth]{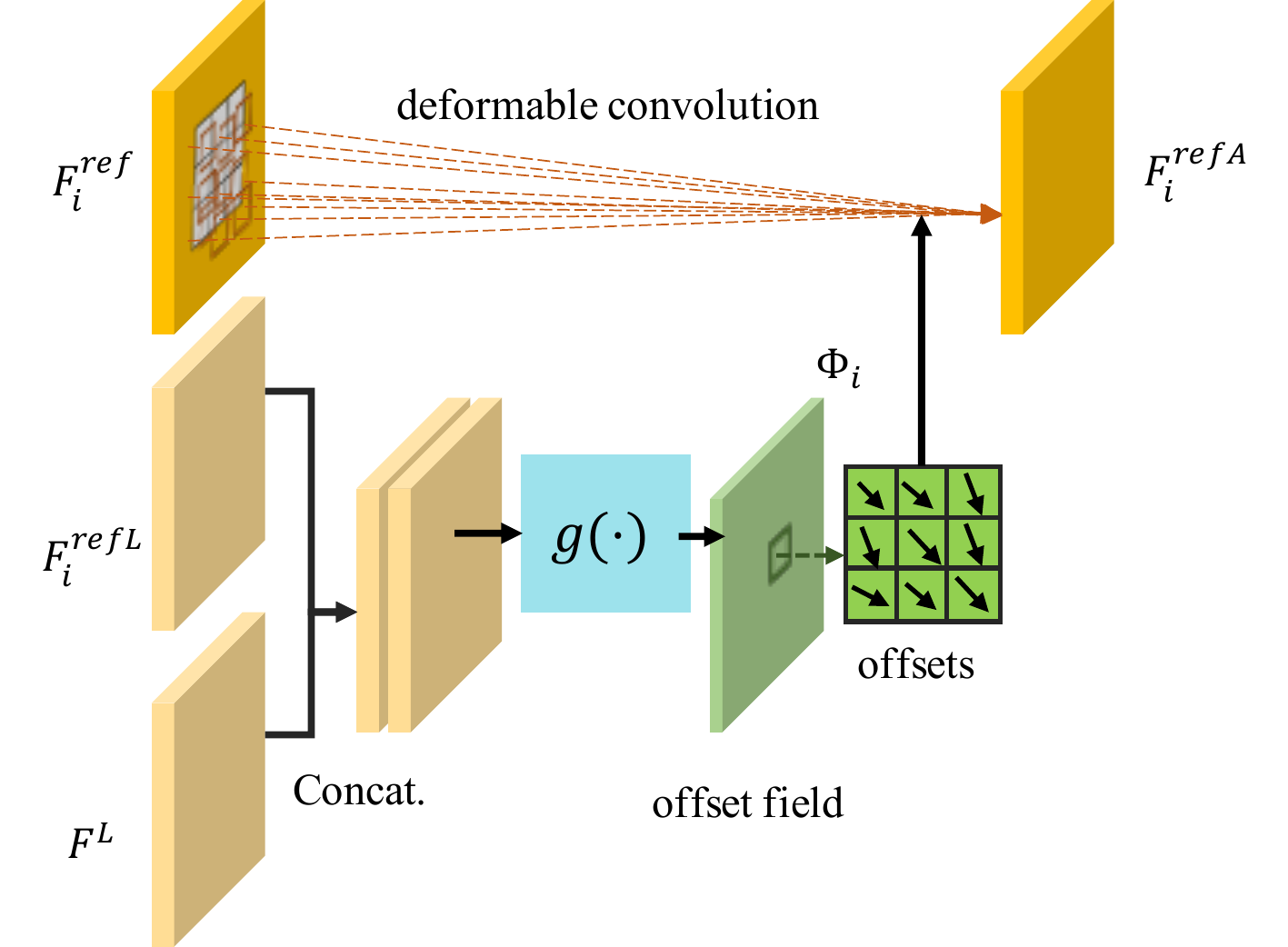}/
  \subcaption{(a) small module. It is purely based on deformable sampling.}
  \end{subfigure}
  
 \begin{subfigure}[b]{\linewidth}
 \includegraphics[width=\linewidth]{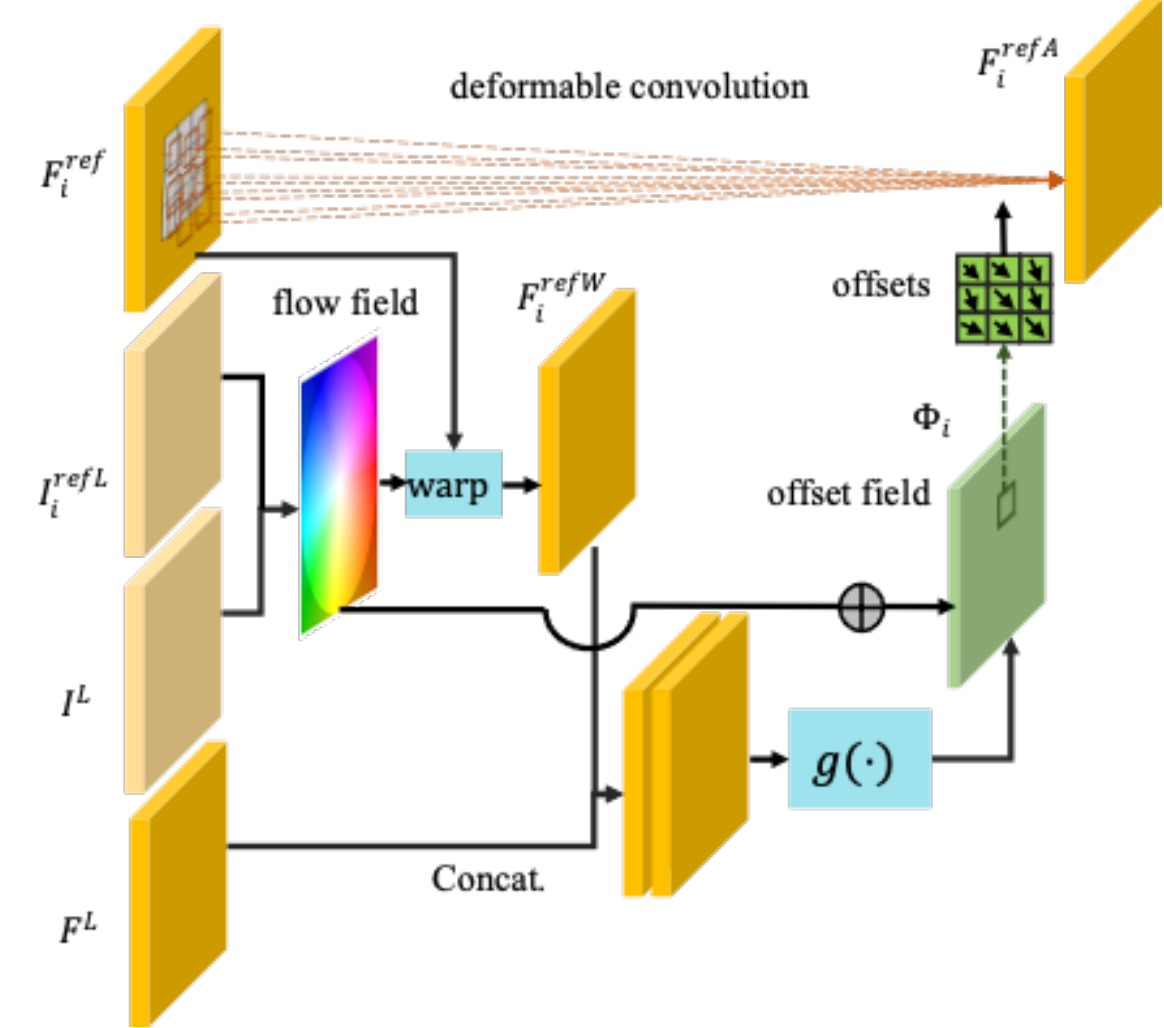}
 \subcaption{(b) large module with optical flow and deformable alignment.}
\end{subfigure}

\end{center}
\caption{Reference feature alignment (RFA) used in small and large model.}
 \label{fig:d_align}
\vspace{-3mm}
\end{figure}

Given extracted feature maps $F^{L}$ from the input LR image and $\{ F^{ref}_i\}_{i=1}^n$ from the reference images, we want to acquire guiding features that are well-aligned with the contents of the LR image to mitigate any mismatches in view or pose. To achieve this goal, We propose learning a feature alignment function $f(\cdot)$ to directly align the reference feature maps $F^{ref}_i$ as shown in Figure~\ref{fig:d_align}. A general form of the alignment function can be formulated as: 
\begin{equation}
    F^{refA}_i = f(F^{ref}_i, I^{refL}_i, I^L, F^L) = T(F_i^{ref}, \Phi_i),
\label{eq:samp_fun}
\end{equation}
where $F^{refA}_i$ denotes the $i$-th aligned reference feature, $T(\cdot)$ is the sampling function, and $\Phi^i$ is the corresponding sampling parameters. Inspired by the deformable alignment~\cite{dai2017deformable,zhu2019deformable} in \cite{wang2019edvr,tian2020tdan,xiang2020zooming} for spatial and temporal super-resolution, we propose to use deformable sampling functions to implicitly capture the similarities between LR content and reference images. However, the training of deformable alignment module is hard and full of instability, which might impair the model's final performance. To overcome this issue, we combine the optical flow as guidance.

The offset for the deformable sampling function should be learned based on the correspondences between the reference image and the input LR image, which is very similar to the goal of optical flow. Thus, we directly merge the optical flow into the offset of deformable alignment, and compute the offset residue to further improve the accuracy. We first estimate the optical flow $o_{i}$ between $I^L$ and $I^{refL}_i$, and use it to warp the reference features:
\begin{equation}
    F_i^{refW} = warp(F^{ref}_i, o_i)
\end{equation}

Then the warped reference feature is used to predict the offset residual $\Delta p_i$, along with the LR feature $F^L$:
\begin{equation}
    \Delta p_i = g([F^{refL}_i,F^{L}]),
\end{equation}
where $g(\cdot)$ denotes a general operation of convolution layers for the offset estimation; $[\cdot,\cdot]$ denotes channel-wise concatenation. Then we can acquire the sampling parameters $\Phi_i=o_i + \Delta p_i$. With the flow-guided offset, the sampling function in Equation~\ref{eq:samp_fun} can be performed with a deformable convolution~\cite{dai2017deformable,zhu2019deformable}:

\begin{equation}
    F^{refA}_i = T(F_i^{ref}, \Phi_i) = DConv(F_i^{ref}, \Phi_i).
\label{eq:align_ref}
\end{equation}

Besides, we construct the HIME (small) by removing the optical flow network in RFA and directly estimating the offset without flow guidance, as shown in Fig.~\ref{fig:d_align} (a).

\subsection{Content-conditioned Feature Aggregation}
\label{subsec:cofa_module}
\begin{figure}[t]
    \centering
    \includegraphics[width=.95\linewidth]{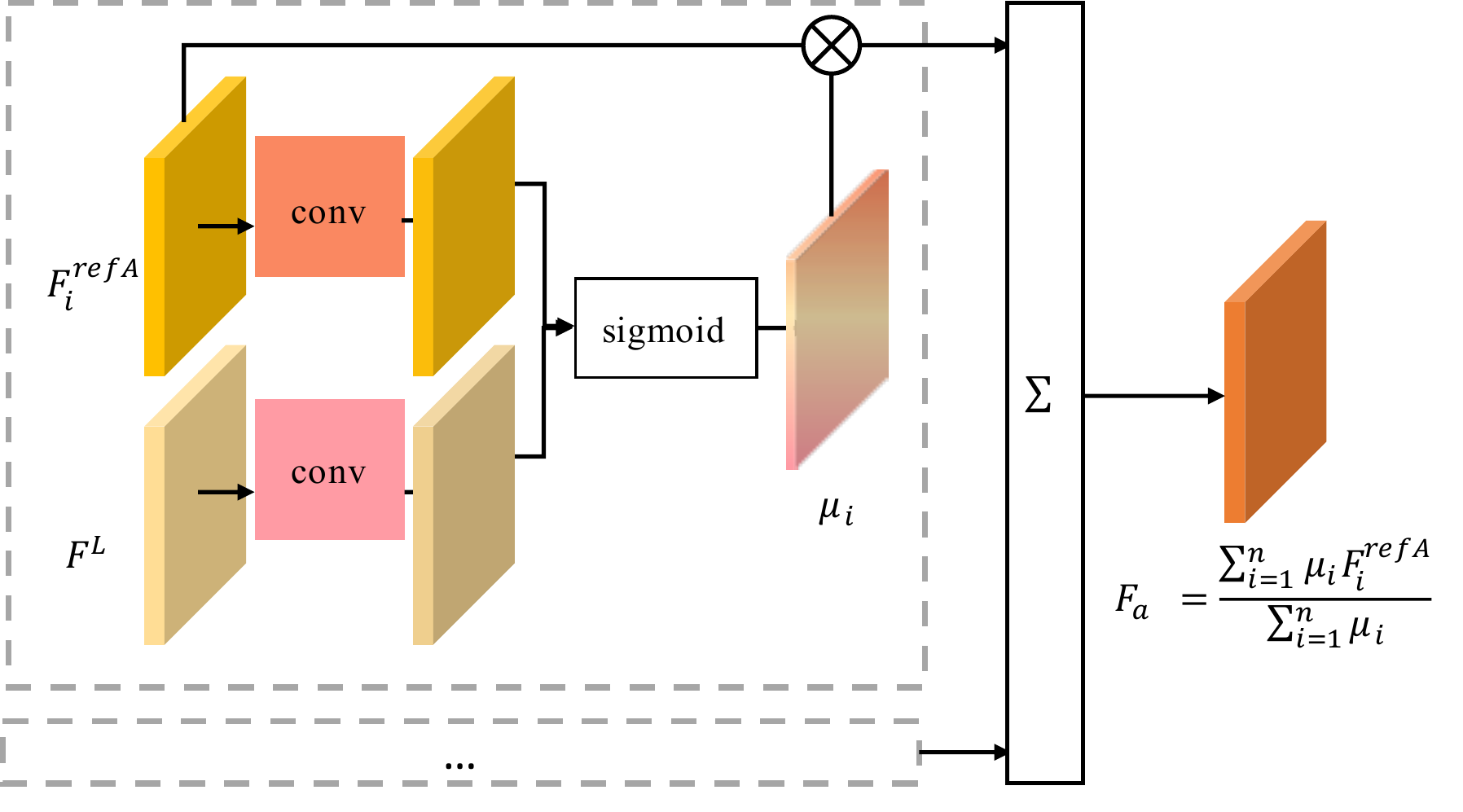}
    \caption{Content-conditioned feature aggregation (CoFA): For each aligned reference feature, we compute a similarity score $\mu$ and then aggregate all features with a weighted average. }
    \label{fig:ssaf}
    \vspace{-3mm}
\end{figure}

Now we have a set of aligned reference feature maps: $\{F^{refA}_i\}_{i=1}^n$ for the following feature transferring and reconstruction steps. As shown in Figure~\ref{fig:ssaf}, the CoFA module aims to map this feature map set to a representation with fixed dimension. In this way, the reference image set with a different number of images can be represented in a unified manner. The representation is determined by all items in the set and conditioned on the LR content. Therefore it can be denoted as:$F_a = \mathcal{F}(F^{refA}_1, F^{refA}_2,\ldots, F^{refA}_n |F^L)$,
where $\mathcal{F}(\cdot)$ is the aggregation function that maps an arbitrary-sized set to a representation of fixed dimension. 

It is challenging to find a proper $\mathcal{F}(\cdot)$ that aggregates features from the whole reference set to obtain an optimized representation. Based on the intuition that references with higher similarity and quality should contribute more to feature transfer, while faces with mismatched features and low-quality features should have less effect on the set representation, we denote $\mathcal{F}(\cdot)$ as:
\begin{equation}
    \mathcal{F}(F^{refA}_1,\ldots, F^{refA}_n |F^L) = \dfrac{\sum_{i=1}^n \mu_i F^{refA}_i}{\sum_i^n \mu_i},
\end{equation}
\begin{equation}
    \mu_i = \mathcal{S} (F^{refA}_i, F^L),
\end{equation}
where $\mathcal{S}(\cdot)$ generates a similarity score $\mu_i$ for the aligned reference feature map $F^{refA}_i$ that is acquired in the same manner as shown by Equation~\ref{eq:align_ref}. Therefore, the final representation of the set is a fusion of each feature weighted by its similarity score. For each aligned reference feature $F^{refA}_i$, the pixel-wise similarity score is calculated as:
\begin{equation}
    \mathcal{S} (F^{refA}_i, F^L) = \sigma(g_1(F^{refA})^T g_2(F^L)),
\end{equation}
where $\sigma(\cdot)$ is sigmoid function that is used for bounding the outputs to the range $[0,1]$ and stabilizing the gradient propagation; and $g_1(\cdot)$ and $g_2(\cdot)$ denotes general convolution layers. The similarity score can also be regarded as an attention mask conditioned on the input content.

Finally, the summation $F_a$ and LR feature map is sent to HR image reconstruction: $F_f = F_a+F^L$. The similarity computation and weighted aggregation steps are parameter-free. Thus, the CoFA module is light-weighted by design.

\subsection{High-Resolution Image Reconstruction}
\label{subsec:hr_recon}
The HR reconstruction module takes the fused feature $F_f$ as input and generates the residual of our target HR output. It is composed of $k_r$ stacked residual blocks~\cite{he2016deep} for learning deep features and a sub-pixel upsampling module with PixelShuffle~\cite{guizar2008efficient} initialized using the ICNR method as in \cite{aitken2017checkerboard,xiang2020boosting}. To encourage the network to focus on learning high-frequency information that is not present in the LR input, we introduce a long-range skip connection to form the final SR output: $ I^{SR} = I^{L}\uparrow_s+\mathcal{R}(F_f),$
where $\uparrow$ denotes the bicubic upscaling operation and $s$ denotes the scale factor; $\mathcal{R(\cdot)}$ denotes the reconstruction operations as described above. Allowing the low-frequency information in the LR input to bypass the reconstruction network lowers the difficulty of reconstruction learning and accelerates the convergence of the optimization process. We use the Charbonnier penalty function~\cite{lai2017deep} as the loss term for pixel-wise reconstruction to optimize our framework: $L_{rec} = \sqrt{||I^{HR} - I^{SR} ||^2 + \epsilon^2}$,
where $I^{HR}$ denotes the ground-truth HR frame, and $\epsilon$ is empirically set to $1\times10^{-3}$. 

Since the input and reference images are highly related in the face domain, our model can simultaneously learn the feature alignment and similarity score with only supervision from the HR ground truths through the end-to-end training.

\section{Correlation Loss}
\label{sec:cor_loss}
\noindent \textbf{Motivation.}
The commonly used pixel-wise reconstruction losses inevitably lead to over-smoothing of outputs and don't match the human visual perception of natural images~\cite{lai2017deep}, since they fail to capture the underlying local relationships between pixels. While the perceptual loss~\cite{simonyan2014very} and style loss~\cite{gatys2016image} have been introduced to provide more perception-oriented supervision, they require a pretrained network from another high-level vision task, and are not versatile for representing textures of very high-resolution images due to the limits of training data. To effectively represent the local texture patterns of different scales in a controllable manner, we devise the correlation loss. It first builds a correlation map from the correlation between the center pixel and its neighbors to represent the spatial patterns. Thus, matching the correlation map can help the network reconstruct more realistic details and improve the perceptual quality of the output images.

\begin{figure}[t]
    \centering
    \includegraphics[width=\linewidth]{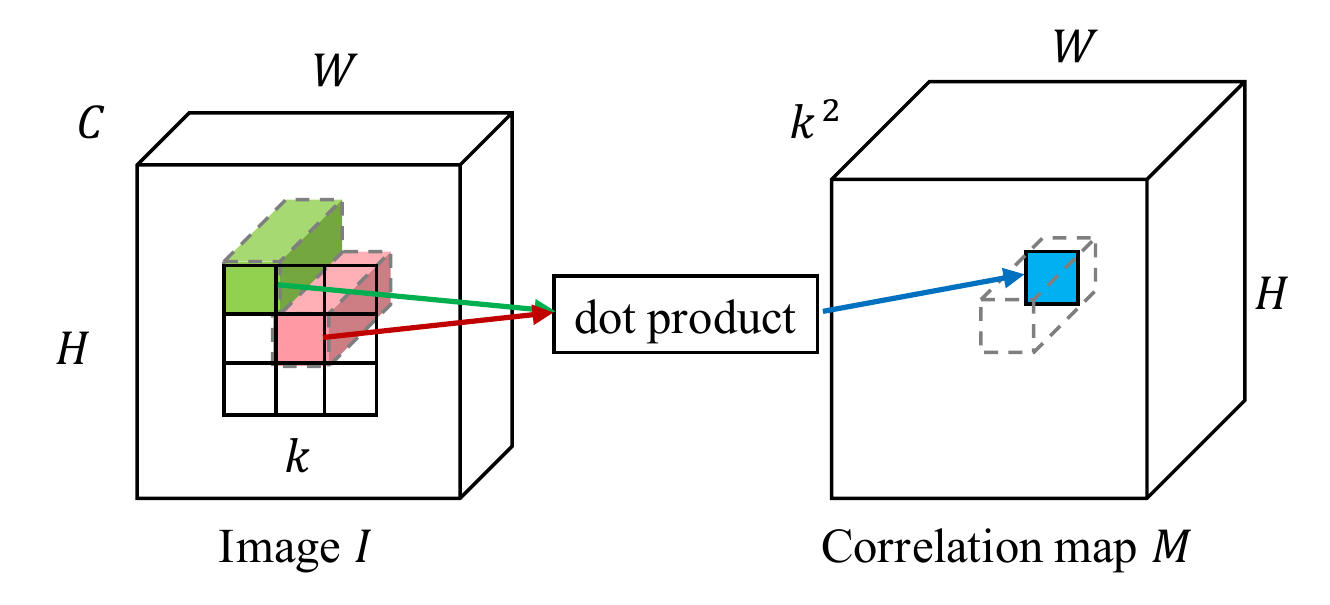}
    \vspace{-2mm}
    \caption{Illustration of the proposed correlation loss. The correlation operator is used for both generated and ground-truth images. Then we take the corresponding output correlation maps to calculate the correlation loss.}
    \label{fig:correlation}
    \vspace{-3mm}
\end{figure}

\vspace{1mm}
\noindent \textbf{Design of Correlation Loss.}
As shown in Figure~\ref{fig:correlation}, each image $I$ can be represented by a 3D tensor of size $(C, H, W)$, where $C$ is the number of channels and $(H, W)$ denotes the spatial resolution. We first subtract the mean of each channel to center the data around 0. For a given pixel $I(x,y)$, we calculate its inner product with the neighboring pixels $I(x-i, y-j)$ as well as itself within a $k\times k$ window:
\begin{equation}
    cor(i,j,x,y) = \dfrac{1}{k^2}\langle I(x,y), I(x-i, y-j)\rangle,
\end{equation}
where $\langle,\rangle$ denotes inner product, $i,j\in \lfloor -\dfrac{k+1}{2}, \dfrac{k+1}{2} \rfloor$, and $\dfrac{1}{k^2}$ is for normalization. $k$ is the maximal displacement for computing the local correlation. 
As a result, we can acquire a correlation map $M_{cor}$ of size $(k\times k, H, W)$. The correlation loss is the distance between the correlation maps from the ground truth HR and the generated SR images:
\begin{equation}
    L_{cor} = ||M_{cor}^{HR} - M_{cor}^{SR}||.
\end{equation}

In our implementation, we adopt the $L1-$distance for this loss term. A larger window size $k$ can encode more information while quadratically increasing the computational cost. Thus, we define the dilated correlation following the same manner as the dilated convolution~\cite{yu2015multi}. By increasing the dilation factor $d$, we can enlarge the correlation window from $k\times k$ to $(kd -d +1)\times (kd -d +1)$.

\begin{figure}[htbp]
\captionsetup[subfigure]{labelformat=empty}
\begin{center}
  \begin{subfigure}[b]{\cocowidth\linewidth}
  \includegraphics[width=\linewidth]{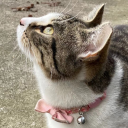}/
  \subcaption{Image}
  \end{subfigure}
  \begin{subfigure}[b]{\cocowidth\linewidth}
  \includegraphics[width=\linewidth]{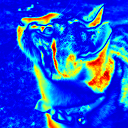}
  \subcaption{$k=3$}
  \end{subfigure}
 \begin{subfigure}[b]{\cocowidth\linewidth}
 \includegraphics[width=\linewidth]{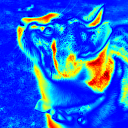}
  \subcaption{$k=5$}
  \end{subfigure}
\begin{subfigure}[b]{\cocowidth\linewidth}
 \includegraphics[width=\linewidth]{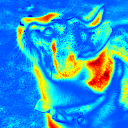}
  \subcaption{$k=7$}
  \end{subfigure}
\end{center}
\caption{Visualization of correlation maps of different window sizes. The image is $128\times 128$-resolution.}
 \label{fig:vis_cor_maps}
\vspace{-3mm}
\end{figure}

\vspace{1mm}
\noindent \textbf{Visualizing Correlation Maps.}
\label{subsec:vis_cor_map}
To better understand the correlation operation, we visualize the correlation maps of the HR image with different correlation kernel window sizes $k\in \{3,5,7\}$. In Figure~\ref{fig:vis_cor_maps}, we observe that the correlation map encodes the original image based on the local textures. In each correlation map, the blue areas correspond to the regions with more high-frequency features, like furs and the background, regardless of the color difference. While the red regions are more smooth, \eg, the brightest and darkest part of the fur. With the increase of window size $k$, the correlation operator perceives and encodes features within a broader area, and thus looks more coarse-grained in the visualized results.

\section{Experiment}
\label{sec:exp}

\subsection{Implementation Details}
\label{subsec:impl}
In our implementation, $k_l = 5$, $k_h = 3$, and $k_r=20$ residual blocks are used in LR feature extraction, HR feature extraction, and HR image reconstruction modules, respectively. For each LR input, we randomly select three different HR images to build the reference set during training. We adopt SpyNet~\cite{ranjan2017optical} as the optical flow estimator in HIME (large). 

\noindent \textbf{Objective Function.} For a fair comparison with previous methods, we train two types of models: reconstruction-oriented models HIME$_{rec}$ with $L_{rec}$ only; and perception-oriented models HIME$_P$ including the pixel-wise reconstruction loss $L_{rec}$, the adversarial loss $L_{adv}$, the perceptual loss $L_{per}$, and our proposed correlation loss $L_{cor}$:
\begin{equation}
    \mathcal{L}_{P} =  \lambda_{rec} L_{rec}+  \lambda_{adv} L_{adv} + \lambda_{per} L_{per} + \lambda_{cor} L_{cor},
\label{eq:loss}
\end{equation}
where $\lambda$s are the weights for each loss term. In our implementation, $\lambda_{rec} = 1.0, \lambda_{adv}=0.1, \lambda_{per}=0.01, \lambda_{cor}=0.1$. The pixel-wise reconstruction loss and the correlation loss are already described in the main paper. For the perceptual loss, we adopt the structure of VGG-19~\cite{simonyan2014very} and extract the features $Fea$ before the ReLU layer. The perceptual loss is measured by $L_1$ distance:
\begin{equation}
    L_{per} = ||Fea^{HR} - Fea^{SR}||_1 .
\end{equation}

We adopt the relativistic GAN~\cite{jolicoeur2018relativistic} for the $L_{adv}$:

\begin{multline}
    L_{adv} = -\mathbb{E}_{HR}[\log(1-D_{Ra}(I^{HR}, I^{SR}))]- \\ \mathbb{E}_{SR}[\log(D_{Ra}(I^{SR}, I^{HR}))],
\end{multline}
where $I^{HR}$ and $I^{SR}$ stand for the ground-truth and generated images, respectively. $D_{Ra}$ denotes the relativistic average discriminator, which can be formulated as:
\begin{equation}
    D_{Ra}(I^{HR}, I^{SR}) = \sigma(C(I^{HR}) - \mathbb{E}_{SR}[C(I^{SR})]),
\end{equation}
where $C(\cdot)$ is the discriminator output, and $\sigma$ is the Sigmoid function, $\mathbb{E}_{SR}[\cdot]$ stands for averaging all $I^{SR}$ in a minibatch. The discriminator loss is defined as:
\begin{multline}
    L_D =  -\mathbb{E}_{HR}[\log(D_{Ra}(I^{HR}, I^{SR}))]- \\ \mathbb{E}_{SR}[\log(1-D_{Ra}(I^{SR}, I^{HR}))].
\end{multline}

\noindent \textbf{Datasets.} CelebAMask-HQ is used as the training and evaluation datasets \cite{CelebAMask-HQ}, including over 30,000 high-resolution headshots selected from the CelebA dataset~\cite{liu2015faceattributes}. We acquire the identity information from the original CelebA dataset and remove 3,300 out of 6,217 identities with $< 4$ images, which are not enough to construct a set of multiple references. The remaining identities are randomly split into a training set and an evaluation set, including 2,600 and 287 identities, respectively. We generate images of different scales by bicubic downsampling with factor $=s$. For each LR input, we randomly select three different HR images to build the reference set during training. In the evaluation stage, we randomly select reference images to form an evaluation list. This list is applied to all evaluated methods for a fair comparison. 

\noindent \textbf{Evaluation Metrics.} We adopt the Peak Signal-to-Noise Ratio (PSNR) and Structural Similarity Index (SSIM) \cite{wang2004image} metrics to evaluate the reconstruction performance on all RGB channels. We also compare the perceptual quality with LPIPS~\cite{zhang2018unreasonable}. To measure the efficiency of the different methods, we report the model parameters and computational cost for each setting.

\subsection{Comparison to the State of the Art}
\label{subsec:sota}
We evaluate the performance of our HIME network under the $4 \times$ and $8 \times$ upsampling setting following the previous approaches. For $4 \times$ upscale, we compare two SOTA RefSR methods: SRNTT~\cite{zhang2019image}\footnote{PyTorch implementation: \href{https://github.com/S-aiueo32/srntt-pytorch}{https://github.com/S-aiueo32/srntt-pytorch}} and TTSR~\cite{yang2020learning}, and three recent face restoration method SPARNet~\cite{chen2020learning}, PSFR-GAN~\cite{chen2020progressive} and DFDNet~\cite{li2020blind}. We did not test DFDNet~\cite{li2020blind} on the $32\times 4$ setting since its face and landmark detectors cannot handle such tiny faces. For $8 \times$ upsampling, we compare our method with five face hallucination methods: PFSR~\cite{bulat2018learn}, FSRNet~\cite{chen2018fsrnet}\footnote{PyTorch implementation: \href{https://github.com/cydiachen/FSRNET_pytorch}{https://github.com/cydiachen/FSRNET\_pytorch}}, GWAInet~\cite{dogan2019exemplar}, SPARNet~\cite{chen2020learning} and PSFR-GAN~\cite{chen2020progressive}. Quantitative results are shown in Table~\ref{tab:x4x8_res}.

\begin{table}[htbp]
\begin{center}
\caption{Quantitative comparison of our results and other SOTA methods. The best results are shown in \textbf{bold}.}
\label{tab:x4x8_res}
\vspace{-2mm}
\resizebox{\columnwidth}{!}{
\begin{tabular}{c|cccccc}
\hline
(LR, $s$)          & Methods & PSNR & SSIM & LPIPS & Params (M) & GMACs \\
\hline
\multirow{3}{*}{$(32, 4)$} & Bicubic &   25.64   &   0.7752   &   0.3229    &    -        &   -    \\
                    & SRNTT~\cite{zhang2019image}   &  28.02    &   0.8434   &   0.0682    &    6.30        &   36.47    \\
                    & TTSR~\cite{yang2020learning}    &  27.31    &  0.8346    &    0.0633   &     6.73       &   26.62    \\ 
                    & SPARNet~\cite{chen2020learning}    &  20.50    &  0.6118    &   0.1617   &     85.73       &   45.25    \\ 
                    & PSFR-GAN~\cite{chen2020progressive}    &  25.47    &  0.7709    &   0.0981   &     67.05       &   117.84    \\ 
                    \cline{2-7}
                    & HIME$_{rec}$ (small)    &   29.11   &  0.8794    &   0.1136    &      \textbf{0.87}      &   \textbf{1.86}    \\
                    & HIME$_P$ (small)    &    27.16           &   0.8269   & 0.0464 & \textbf{0.87}   &   \textbf{1.86}    \\
                    & HIME$_{rec}$ (large)   &   \textbf{29.23}   &  \textbf{0.8817}    &   0.1102    &     9.23      &   6.06    \\
                    & HIME$_P$ (large)   &    27.05           &   0.8224   & \textbf{0.0461} & 9.23   &   6.06    \\
                    \hline
\multirow{3}{*}{$(64, 4)$} & Bicubic &  28.40    &    0.8169  &    0.2860   &      -      &    -   \\
                    & SRNTT~\cite{zhang2019image}   &    30.41  &   0.8552   &   0.0906    &     6.30       &    145.89   \\
                    & TTSR~\cite{yang2020learning}    &   29.87   &   0.8484   &   0.0851    &   6.73         &   106.48    \\ 
                    & SPARNet~\cite{chen2020learning}    &  23.26    &  0.6990    &   0.1341   &     85.73       &   180.99    \\ 
                    & PSFR-GAN~\cite{chen2020progressive}    &  26.62    &  0.7685    &   0.1039   &     67.05       &   161.89    \\ 
                    & DFDNet~\cite{li2020blind}    &  21.55    &  0.6587    &  0.1581    &  133.34 &   601.04    \\ 
                    \cline{2-7}
                    & HIME$_{rec}$ (small)    &   31.24   &  0.8785    &    0.1611   &       \textbf{0.87}     &   \textbf{7.48}    \\
                    & HIME$_P$ (small)    &       29.06        & 0.8262      & \textbf{0.0633} & \textbf{0.87}   &   \textbf{7.48}    \\
                    & HIME$_{rec}$ (large)    &   \textbf{31.28}   &  \textbf{0.8789}    &    0.1600   &      9.23     &  24.24    \\
                    & HIME$_P$ (large)    &       29.16        & 0.8272      & 0.0641 & 9.23   &   24.24    \\
\hline
\multirow{3}{*}{$(16, 8)$} & Bicubic &   21.83   &   0.5929   &  0.5247     &    -        &    -   \\
               &   PFSR\cite{bulat2018learn}    &   21.44   &   0.5778   &   0.2065    &    10.08        &   8.97    \\
& FSRNet~\cite{chen2018fsrnet}  &   20.03   &  0.5749    &    0.2865   &      15.52      &    3.20   \\
& GWAINet~\cite{dogan2019exemplar} &   21.96   &  0.5844    &    0.2056   &    4.29        &   6.55    \\
& SPARNet~\cite{chen2020learning}    &  19.00    &  0.5022    &   0.2576   &     85.73       &   45.25    \\ 
& PSFR-GAN~\cite{chen2020progressive}    &  22.05    &  0.6102    &  0.2062   &     67.05       &   117.84    \\ 
                    \cline{2-7}
         &           HIME$_{rec}$ (small)    &  24.54    &   0.7411   &   0.2433    &     \textbf{0.90}       &    \textbf{0.49}   \\
& HIME$_{P}$ (small)    &     22.45 &   0.6338   &   \textbf{0.1297}    &     \textbf{0.90}       &    \textbf{0.49}   \\
         &           HIME$_{rec}$ (large)    &  \textbf{24.68}    &   \textbf{0.7467}   &   0.2361    &     9.26       &    4.49   \\
& HIME$_{P}$ (large)    &   23.35   &    0.6744  &     0.1313  &      9.26      &    4.49   \\
\hline
\end{tabular}
}
\end{center}
\vspace{-3mm}
\end{table}

\newcommand{\qualiwidth}{0.12} 
\begin{figure*}[htbp]
\captionsetup[subfigure]{labelformat=empty}
\begin{center}
  \begin{subfigure}[b]{\qualiwidth\linewidth}
  \includegraphics[width=0.9\linewidth]{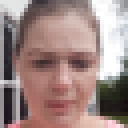}
  \end{subfigure}
\begin{subfigure}[b]{\qualiwidth\linewidth}
 \includegraphics[width=0.9\linewidth]{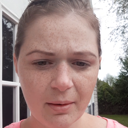}
  \end{subfigure}
  \begin{subfigure}[b]{\qualiwidth\linewidth}
  \includegraphics[width=0.9\linewidth]{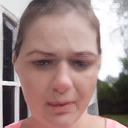}
  \end{subfigure}
  \begin{subfigure}[b]{\qualiwidth\linewidth}
 \includegraphics[width=0.9\linewidth]{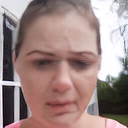}
  \end{subfigure}  
  \begin{subfigure}[b]{\qualiwidth\linewidth}
  \includegraphics[width=0.9\linewidth]{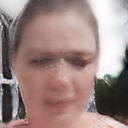}
  \end{subfigure}
  \begin{subfigure}[b]{0.125\linewidth}
 \includegraphics[width=0.864\linewidth]{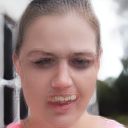}
  \end{subfigure}
   \begin{subfigure}[b]{\qualiwidth\linewidth}
 \includegraphics[width=0.9\linewidth]{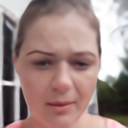}
  \end{subfigure}
\begin{subfigure}[b]{\qualiwidth\linewidth}
 \includegraphics[width=0.9\linewidth]{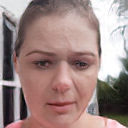}
  \end{subfigure}

  \begin{subfigure}[b]{\qualiwidth\linewidth}
  \includegraphics[width=0.9\linewidth]{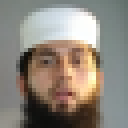}
  \end{subfigure}
\begin{subfigure}[b]{\qualiwidth\linewidth}
 \includegraphics[width=0.9\linewidth]{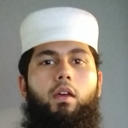}
  \end{subfigure}
  \begin{subfigure}[b]{\qualiwidth\linewidth}
  \includegraphics[width=0.9\linewidth]{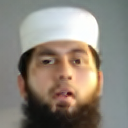}
  \end{subfigure}
  \begin{subfigure}[b]{\qualiwidth\linewidth}
 \includegraphics[width=0.9\linewidth]{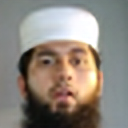}
  \end{subfigure}  
  \begin{subfigure}[b]{\qualiwidth\linewidth}
  \includegraphics[width=0.9\linewidth]{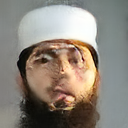}
  \end{subfigure}
  \begin{subfigure}[b]{0.125\linewidth}
 \includegraphics[width=0.864\linewidth]{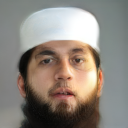}
  \end{subfigure}
   \begin{subfigure}[b]{\qualiwidth\linewidth}
 \includegraphics[width=0.9\linewidth]{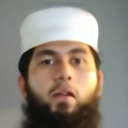}
  \end{subfigure}
\begin{subfigure}[b]{\qualiwidth\linewidth}
 \includegraphics[width=0.9\linewidth]{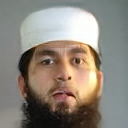}
  \end{subfigure}
  
\begin{subfigure}[b]{\qualiwidth\linewidth}
  \includegraphics[width=0.9\linewidth]{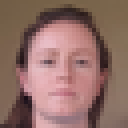}
  \subcaption{Input}
  \end{subfigure}
\begin{subfigure}[b]{\qualiwidth\linewidth}
 \includegraphics[width=0.9\linewidth]{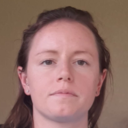}
 \subcaption{GT}
  \end{subfigure}
  \begin{subfigure}[b]{\qualiwidth\linewidth}
  \includegraphics[width=0.9\linewidth]{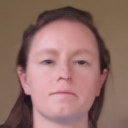}
  \subcaption{SRNTT~\cite{zhang2019image}}
  \end{subfigure}
  \begin{subfigure}[b]{\qualiwidth\linewidth}
 \includegraphics[width=0.9\linewidth]{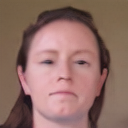}
 \subcaption{TTSR~\cite{yang2020learning}}
  \end{subfigure}  
  \begin{subfigure}[b]{\qualiwidth\linewidth}
  \includegraphics[width=0.9\linewidth]{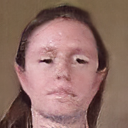}
  \subcaption{SPARNet~\cite{chen2020learning}}
  \end{subfigure}
  \begin{subfigure}[b]{0.125\linewidth}
 \includegraphics[width=0.864\linewidth]{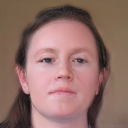}
 \subcaption{PFSR-GAN~\cite{chen2020progressive}}
  \end{subfigure}
   \begin{subfigure}[b]{\qualiwidth\linewidth}
 \includegraphics[width=0.9\linewidth]{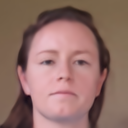}
 \subcaption{HIME$_{rec}$}
  \end{subfigure}
\begin{subfigure}[b]{\qualiwidth\linewidth}
 \includegraphics[width=0.9\linewidth]{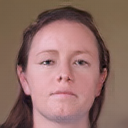}
 \subcaption{HIME$_P$}
  \end{subfigure}
\end{center}
\vspace{-3mm}
\caption{Qualitative comparison with SOTA methods for $4 \times$ upscale setting. Input resolution: $32\times 32$.}
 \label{fig:x4_res}
\vspace{-2mm}
\end{figure*}

From Table~\ref{tab:x4x8_res}, we can learn the following facts: (1) reference-based SR methods, like SRNTT, TTSR and our HIME, demonstrate better performance than other non-reference approaches on both distortion-oriented metrics and perception-oriented metrics, which validate that using references can improve the SR fidelity. Our network outperforms the other result by 1.21/1.09 dB on $(32, 4)$, and 0.87/0.83 dB on $(64, 4)$; (2) Although SRNTT and TTSR have fewer parameters than other compared methods, their computational costs are relatively high due to the exhaustive search during feature matching. With the learnable feature extractors, our small model is over $7\times$ smaller than SRNTT and TTSR. The reference feature alignment in LR space makes our network have $14.3$ and $4.39$ $\times$ fewer GMACs than TTSR. For the $(16, 8)$ setting, we can observe that our method performs well even under the very challenging $8 \times$ upsampling setting. 

The visual results on the DFDC dataset~\cite{dolhansky2020deepfake} are shown in Figure~\ref{fig:x4_res}, which validates our observations above. RefSR methods like SRNTT, TTSR, and ours can generate more robust and visually pleasing results. For the GAN-based face enhancement methods SPARNet and PFSR-GAN, while their results are rich in details, sometimes they fail on tiny faces with deformations. 

\subsection{Ablation Study}
\label{subsec:abl}
We perform a comprehensive ablation study to further demonstrate the effectiveness of different modules in our network and the correlation loss. We also discuss the influence of face chirality under different data augmentation strategies. All experiments below are conducted under the same setting: $8 \times$ upscale with input size $16 \times 16$ images.

\vspace{1mm}
\noindent \textbf{Effectiveness of Reference Feature Alignment.} To investigate the proposed RFA module, we compare three models: (a), (b), and (c), where (a) replaces the deformable convolution in the RFA module with common convolution that does not have the capability of feature alignment, and (b) is our small model by removing the optical flow guidance, and directly estimate the offset with $F^{ref}_i$ and $F^L$, (c) is our large model as illustrated in Section~\ref{subsec:rfa_module}

\begin{table}[htbp]
\begin{center}
\caption{Ablation study of feature alignment methods.}
\label{tab:abl:RFA}
\vspace{-2mm}
\resizebox{0.7\columnwidth}{!}{
\begin{tabular}{cccc}
\hline
Methods & PSNR$\uparrow$ & SSIM$\uparrow$  & LPIPS$\downarrow$ \\
\hline
Conv &  24.33    &   0.7311   &   0.2605      \\
Dconv  &  24.54    &  0.7411    &   0.2433       \\
Dconv-flow  &  24.68    &  0.7467    &    0.2361     \\
\hline
\end{tabular}
}
\end{center}
\vspace{-3mm}
\end{table}

From Table \ref{tab:abl:RFA}, we can see that adopting the deformable alignment brings up the performance on all metrics compared with using the common convolution. And the flow-guided deformable alignment can further improve the performance. The results demonstrate that our RFA module can better match the features between the LR content and the references and is more robust to the misalignment and distortion. Our network conducts the offset computation and feature matching in the LR space, achieving a better performance while reducing the computational cost.

\noindent \textbf{Set Feature Aggregation.} To validate the effect of our proposed feature aggregation mechanism in the CoFA module, we compare three different models: (a) averages the features without content conditioning, (b) aggregates the features by max-pooling across the set, and (c) is our proposed aggregation method weighted by the learned content similarity. The quantitative results are shown in the Tabel~\ref{tab:abl:cofa}.

\begin{table}[htbp]
\begin{center}
\caption{Ablation study of feature aggregation methods.}
\label{tab:abl:cofa}
\vspace{-2mm}
\resizebox{0.7\columnwidth}{!}{
\begin{tabular}{cccc}
\hline
Methods & PSNR$\uparrow$ & SSIM$\uparrow$  & LPIPS$\downarrow$ \\
\hline
Average &    22.120  &   0.6350   &   0.4332      \\
Max-pool    &   22.118   &  0.6349    &   0.4331       \\
CoFA  &   24.381   &  0.7339    &   0.2533      \\
\hline
\end{tabular}
}
\end{center}
\vspace{-3mm}
\end{table}

From Table~\ref{tab:abl:cofa}, we can see that the model with our content-conditioned feature aggregation module outperforms the average and max pooling by over 2 dB in terms of PSNR. Adopting the CoFA module greatly improves performance on all metrics, which indicates that our designed module can extract a better set representation, helping to restore the LR information and enhance the output quality.

\noindent \textbf{Effect of Multiple Exemplars.} To validate whether using an exemplar set can improve the face super-resolution result, we conduct the following experiments: (a) non-ref: a baseline SR network without references and removing the HR matching and aggregation modules, (b) training and testing with one reference image and (c) with three reference images. From the results in Table~\ref{tab:abl:n_refs}, we can observe that using references significantly increases the PSNR by 0.49 dB while using multiple references further improves it by 0.19 dB. Such improvements also apply to the SSIM and LPIPS. These results verify that our model can benefit from the rich information in the exemplar set, and can effectively utilize the corresponding features to improve the output quality.

\begin{table}[htbp]
\begin{center}
\caption{Ablation study of multiple exemplars by changing the number of references during training and testing.}
\vspace{-2mm}
\label{tab:abl:n_refs}
\resizebox{0.7\columnwidth}{!}{
\begin{tabular}{cccc}
\hline
Num of Ref & PSNR$\uparrow$ & SSIM$\uparrow$  & LPIPS$\downarrow$ \\
\hline
0 &    23.84 &   0.7088  &   0.3440      \\
1    &   24.35  &  0.7318    &  0.2572       \\
3  &  24.54   & 0.7409   &   0.2433      \\
\hline
\end{tabular}
}
\end{center}
\vspace{-3mm}
\end{table}

\textbf{Influence of Reference Images} Our method has the potential to be applied on video calling, where the close-to-camera headshots can be used to enhance the far-away ones when zooming in. For this scenario, we recorded several video calls from ourselves and collected over 5,000 frames to verify the influence of the temporal gap. We downsample these frames $4\times$ to construct LR inputs, and pick an HR image as Ref every $j$ frame. Intuitively, with the increase of interval $j$, the Ref is less similar to the LR inputs due to the motion in natural videos. We also experiment on using a blank image as a reference, which does not provide any similar features. From Table~\ref{tab:vid-temp}, we can observe that the performance decreases with the larger temporal interval and less similarity, and gracefully descends to a lower bound. Still, using Refs shows better results than blank Ref in terms of PSNR and SSIM.

\begin{table}[htbp]
\begin{center}
\caption{Influence of temporal gap between input and reference images.}
\label{tab:vid-temp}
\vspace{-2mm}
\begin{tabular}{ccc}
\hline
Interval $j$    & PSNR & SSIM   \\ \hline
30       & 37.34 & 0.9250 \\ 
60       &  37.24 & 0.9241 \\
120      &  37.12 & 0.9227\\ 
Blank Ref &  36.81 & 0.9207\\
\hline
\end{tabular}
\end{center}
\end{table}

\noindent \textbf{Effect of Correlation Loss.} To justify the effectiveness of correlation loss, we experimentally compare different configurations of HIME in Table~\ref{tab:abl:cor_loss}. We consider the following models: (a) reconstruction loss only; (b) reconstruction loss + correlation loss; (c) multiple losses in GAN training (without correlation loss); (d) correlation loss + (c).

\begin{table}[tbp]
\begin{center}
\caption{Effectiveness of our proposed correlation loss.}
\label{tab:abl:cor_loss}
\vspace{-2mm}
\resizebox{0.7\columnwidth}{!}{
\begin{tabular}{cccc}
\hline
Methods & PSNR$\uparrow$ & SSIM$\uparrow$  & LPIPS$\downarrow$ \\
\hline
$L_{rec}$ &  24.38    &  0.7339    &   0.2533      \\
$L_{rec}+L_{cor}$    &  24.35   &   0.7346    &     0.2437     \\
$L_P$ w/o $L_{cor}$&    22.44  & 0.6204     &     0.1543    \\
$L_P$ w/ $L_{cor}$  &   23.28   &  0.6673    &   0.1389      \\
\hline
\end{tabular}
}
\end{center}
\vspace{-3mm}
\end{table}

From Table~\ref{tab:abl:cor_loss}, by comparing the first two rows, we can observe that introducing the correlation loss slightly decreases the PSNR. However, it improves the structural and perceptual metrics SSIM and LPIPS, which demonstrates that the proposed correlation loss benefits the reconstruction of local textures. Comparing the last two rows, training with the correlation loss greatly leverages the perception-oriented model's performance on all metrics, which further validates the effectiveness of the correlation loss as perception-oriented supervision.

\begin{figure}[tbp]
\captionsetup[subfigure]{labelformat=empty}
\begin{center}
  \begin{subfigure}[b]{0.32\linewidth}
  \includegraphics[width=\linewidth]{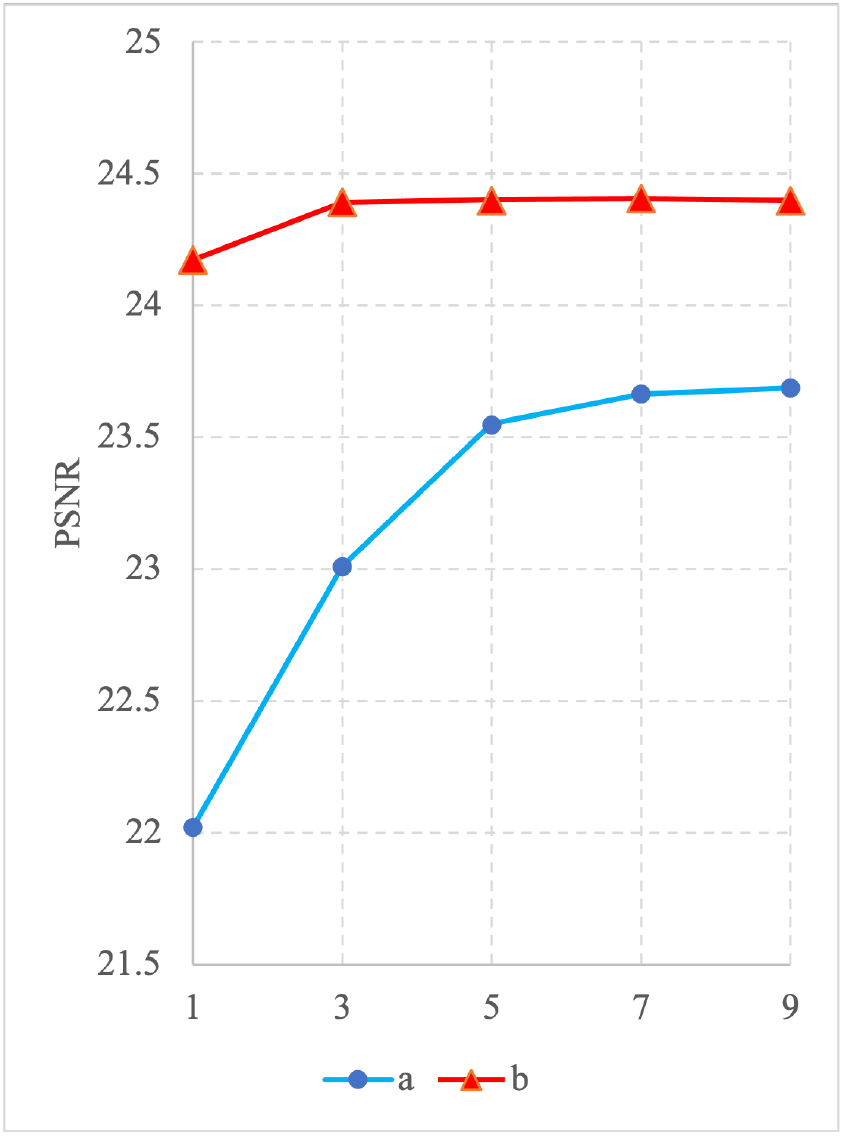}
  \caption{PSNR}
  \end{subfigure}  
  \begin{subfigure}[b]{0.32\linewidth}
  \includegraphics[width=\linewidth]{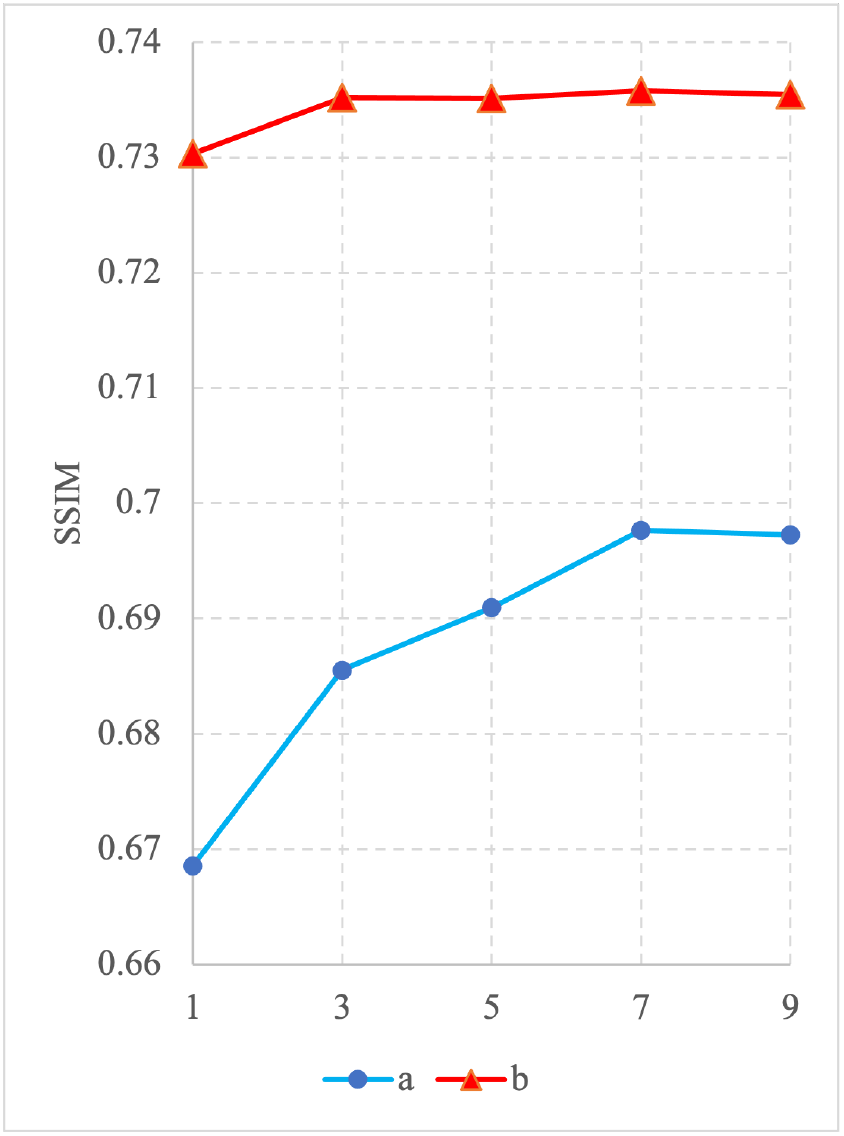}
  \caption{SSIM}
  \end{subfigure}
  \begin{subfigure}[b]{0.32\linewidth}
  \includegraphics[width=\linewidth]{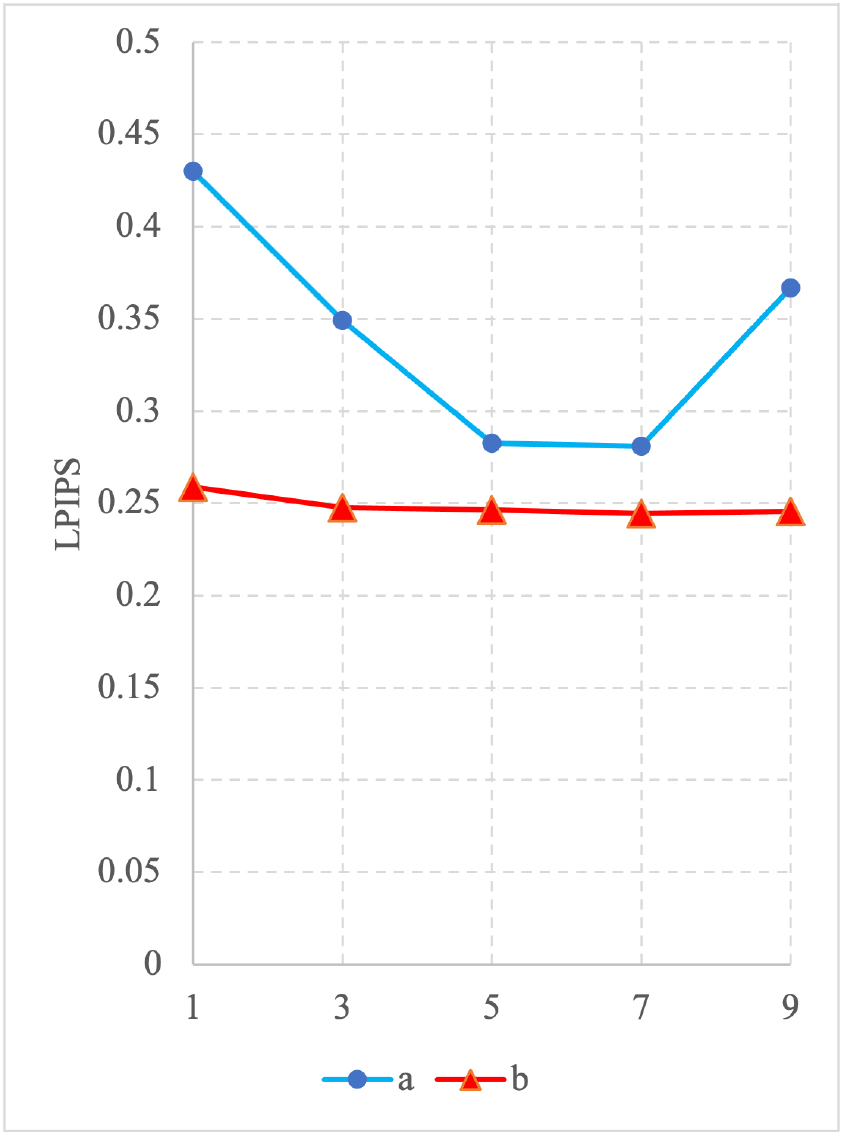}
  \caption{LPIPS}
  \end{subfigure}
\end{center}
\vspace{-2mm}
\caption{Effect of correlation window size $k$ on output quality in terms of PSNR, SSIM, and LPIPS: \textcolor{blue}{(a)} training with $L_{cor}$ only, \textcolor{red}{(b)} fine-tuning with both $L_{rec}$ and $L_{cor}$.}
 \label{fig:cor_loss_size}
\vspace{-3mm}
\end{figure}


Figure~\ref{fig:cor_loss_size} shows the performance of HIME for different correlation window sizes $k\in \{1,3,5,7,9\}$, where $k=1$ degrades to the common $L1$ loss of the squared pixel values. We conduct two types of experiments: (a) training with correlation-loss only (plotted in blue), (b) fine-tuning with both $L_{rec}$ and $L_{cor}$ (plotted in red). Viewing the blue plots, we can observe that with the growth of $k$, the model performs better in terms of PSNR and SSIM. These results demonstrate that the correlation map itself is a good representation of the RGB image. With a larger window size, the correlation map can encode more information. Still, such improvement becomes more marginal when $k$ is large enough. When $k=9$, the LPIPS even increases. As for the red plots, we can see a similar trend: when $k\geq 3$, the improvement on PSNR and SSIM is very trivial. These results indicate that for a certain scale, there exists a range of $k$ that work best in representing the local patterns. Within this range, the LPIPS scores keep decreasing with the increase of $k$. It implies that the correlation loss is more like perception-oriented supervision, which validates our description in Section~\ref{sec:cor_loss}.

\noindent \textbf{Face Chirality.} Typical human faces contain a variety of asymmetries. ~\cite{lin2020visual} brings up the visual chirality in faces and the distribution bias in public face datasets. Here we compare the influence of such asymmetries in headshot RefSR to answer the following questions: 1) Does the mismatch between input LR and references matter? 2) Does the bias in the training set influence the reconstruction performance? 

\begin{table}[htbp]
\begin{center}
\caption{Influence of face chirality.}
\label{tab:abl:chirality}
\vspace{-2mm}
\resizebox{\columnwidth}{!}{
\begin{tabular}{cccc|cc}
\hline
Models & No-aug & Uneven h-flip  & Even h-flip & PSNR$\uparrow$ & SSIM$\uparrow$ \\
\hline
(a) &   $\surd$   &      &     & 22.60  &  0.660  \\
(b)  &      &  $\surd$    &     & 22.52 &  0.654    \\
(c) &      &      &  $\surd$ &  23.63   &  0.662 \\ 
\hline
\end{tabular}
}
\end{center}
\vspace{-3mm}
\end{table}

Tabel~\ref{tab:abl:chirality} shows our experimental results of changing the augmentation: (a) no augmentation; (b) randomly horizontal-flip the LR or Ref images, but not both for a given pair, which introduces the face view mismatch; (c) randomly horizontal-flip both the LR and reference images, which balances the number of left and right faces without introducing mismatches. By comparing (a) and (b), we can observe that the PNSR drops by 0.08 and the SSIM drops by 0.006, respectively, which indicates that training with mismatched views of faces would impair the model's performance. Compared with (a), (c) performs slightly better in terms of the PSNR and SSIM, which demonstrates that the proper augmentation improves the performance by mitigating the distribution bias in the dataset.

\section{Conclusion and Future Work}
\label{sec:conclusion}
In this paper, we propose an effective framework for headshot image super-resolution with multiple exemplars without face structure priors. To achieve this, we introduce a reference feature alignment module to search and align corresponding features to the LR content. To construct an optimized set representation, we propose a feature aggregation network conditioned on the input content. With such a design, our network can learn to fully utilize the rich information in the exemplar set and be robust to misalignment and deformations.
Furthermore, we propose a correlation loss that supervises the reconstruction of local textures with correlation maps. We believe that our new \textbf{H}eadshot \textbf{I}mage Super-Resolution with \textbf{M}ultiple \textbf{E}xemplars network (HIME) provides a novel idea to efficiently utilize a set of data for the reference-based super-resolution and face hallucination task. In future works, we will explore other aggregation methods to generate a better set representation with the aid of face priors. In addition, we will further validate the effectiveness of the correlation loss as generic supervision for other low-level tasks, \eg image denoising, video frame interpolation, style transfer, \etc.

\bibliographystyle{ACM-Reference-Format}
\bibliography{egbib.bib}

\end{document}